# Towards a New Interpretation of Separable Convolutions

Author: Tapabrata Ghosh[1] (admin@Ingemini.com)
Ingemini LLC[1]

**Abstract:**
In recent times, the use of separable convolutions in deep convolutional neural network architectures has been explored. Several researchers, most notably (Chollet, 2016) and (Ghosh, 2017) have used separable convolutions in their deep architectures and have demonstrated state of the art or close to state of the art performance. However, the underlying mechanism of action of separable convolutions are still not fully understood. Although their mathematical definition is well understood as a depthwise convolution followed by a pointwise convolution, "deeper" interpretations such as the "extreme Inception" hypothesis (Chollet, 2016) have failed to provide a thorough explanation of their efficacy. In this paper, we propose a *hybrid* interpretation that we believe is a better model for explaining the efficacy of separable convolutions.

# 1. Introduction

Convolutional neural networks have quickly come to dominate various tasks in artificial intelligence, most notably computer vision tasks such as image classification and object detection. Other computer vision tasks such as face recognition, person re-identification, semantic segmentation and semantic regression have also been successfully tackled, enabling for the first time in history, for computers "to see".

The vision ability of convolutional neural networks has been a direct consequence of the revolution in depth, in which the amount of stacked convolutional layers has been increased dramatically. This strategy of depth was initially introduced in AlexNet (Krizhevsky et al, 2012) and has been continued with VGG and other networks. However, purely increasing depth will not lead to proper training and good results, for this reason Inception modules (Szegedy et al, 2015) and/or residual connections are needed (He et al, 2015). However, in recent times, separable convolutions have gained considerable interest due to their incredible ability to train extremely deep models and achieve state of the art results.

Separable convolutions have recently been utilized for the construction of deep convolutional neural networks. Most notably, their efficacy was highlighted by Chollet,2016 in the XCeption architecture and both their efficiency and efficacy was highlighted by Ghosh, 2017 in the QuickNet architecture. However, a satisfactory explanation as to *why* separable convolutions are able to produce state of the art results has not been found until now. Separable convolutions are defined to be a depthwise convolution followed by a pointwise (1x1) convolution. Notably, the XCeption architecture (Chollet, 2016) used separable convolutions in tandem with residual connections whereas QuickNet does not use residual connections (Ghosh, 2017). The latter work uncovers a peculiar property of separable convolutions in that it appears they are not fully dependent upon the existence of residual connections for training of deep networks.

## 2. Related Work

Our hypothesis relies upon three main works, grouped convolutions(which dates back to Krizhevsky, 2012), ResNeXt (Xie et al, 2016) and XCeption (Chollet, 2016).

Grouped convolutions (which dates back to Krizhevsky, 2012) can be seen as a generalization of depthwise convolutions since the special case of where number of groups = number of incoming channels is equivalent to a depthwise convolution.

ResNeXt (Xie et al, 2016) describes a new type of network architecture which utilizes "aggregated residual transforms, a building block module they call ResNeXt which enables them to achieve state of the art performance on ImageNet.

XCeption (Chollet, 2016) is an architecture which heavily utilizes separable convolutions in order to reach state of the art accuracy. However, more importantly, Chollet proposes the "Inception hypothesis" which states that separable convolutions can be interpreted as "extreme Inception modules" in where the number of convolution pathways is equivalent to the number of channels. However, we believe that the grouped convolution/depthwise convolution is better explained as a hybrid of ResNeXt and Inception.

## 3. The Hybrid Hypothesis

Chollet proposed in 2016 that separable convolutions can be interpreted as an "extreme version" of an Inception module (Szegedy et al, 2015). Chollet proposed this on the basis of the idea that each channel could be thought of as a new "tower". Unfortunately, this interpretation is very unsatisfying as it ignores the differing filter sizes and receptive fields in each Inception tower. The "Inception hypothesis" proposes that separable convolutions are similar to Inception modules due to the separation of spatial feature extraction and cross-channel feature extraction (through the use of the 1x1 convolution and preservation of cross-channel features through the preceding depthwise convolution). However, we argue that a more befitting hypothesis is a *hybrid* hypothesis between the Inception hypothesis and the ResNeXt (Xie et al, 2016) architecture.

We propose that separable convolutions are best interpreted as a combination of ResNeXt with a cardinality equivalent to the number of channels (depthwise convolutions) and Inception with 1x1 convolutions which analyze cross-channel correlations (the pointwise convolution which follows it). The summation across channels in the pointwise convolution is similarly reminiscent of ResNeXt if we remove the identity connection. Using separable convolutions in coordination with a residual connection allows us to arrive at the "hybrid" equivalence/interpretation to ResNeXt.

      We independently rediscover the conclusions reached by Xie et al in 2016 in which they establish the equivalence of ResNeXt modules to grouped convolutions with cardinality = to the number of incoming channels. Unlike Inception modules, ResNeXt modules are homogeneous and use summation as their model of tensor joining. However, ResNeXt alone is not enough to explain the efficacy of separable convolutions since the crucial cross-channel factor is missing. The use of standard convolutional layers in ResNeXt destroys the ability to efficiently extract cross-channel features. By contrast, the "Inception hypothesis" reasons that separable convolutions preserve the cross-channel correlations and analyze them with the 1x1 convolution. We believe that combining the two paints a more accurate picture of the mechanism of action of separable convolutions.

      Our hypothesis proposes that separable convolutions can achieve their incredible efficacy due to them sharing the advantages of two otherwise state-of-the-art network modules: Inception and ResNeXt. Specifically, we propose that separable convolutions incorporate the advantages of ResNeXt modules due to depthwise convolutions being equivalent to grouped convolutions with the number of groups equal to the number of incoming channels. This is also subsequently thematically equivalent to a ResNeXt block with cardinality equal to the number of incoming channels. Separable convolutions incorporate the advantage of Inception (Szegedy et al, 2015) through the use of the 1x1 convolution which enables separate analysis of cross-channel features. We note that a key difference is the channel-wise feature preservation of the depthwise convolution, in contrast to ResNeXt, which necessitates the invocation of the "Inception hypothesis".

      **In summary, the groups attribute of the depthwise convolution provides the advantages of ResNeXt while the 1x1 convolution and the channel-wise feature preservation of the depthwise convolution provides the advantages of the Inception module.** It is with this combination that separable convolutions are able to achieve their incredible performance.

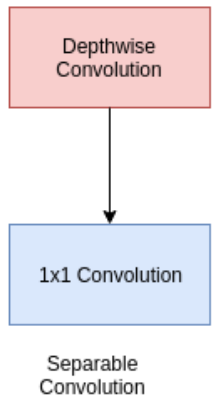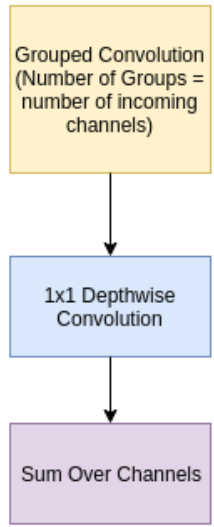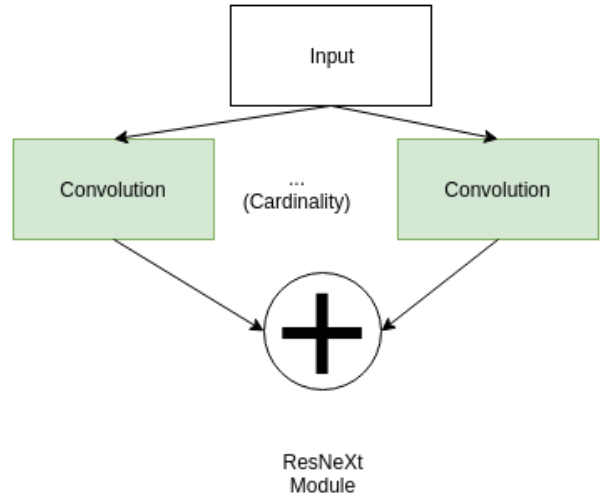

Figure 1

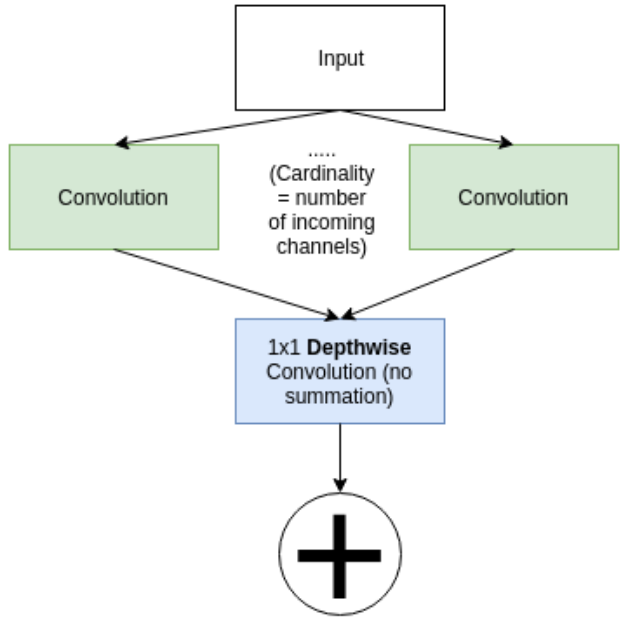 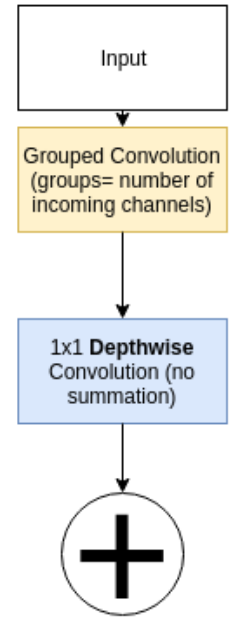

Hypothesized Interpretation of Separable Convolutions

Equivalent Reformulation of the Hypothesized Interpretation of Separable Convolutions

Figure 2

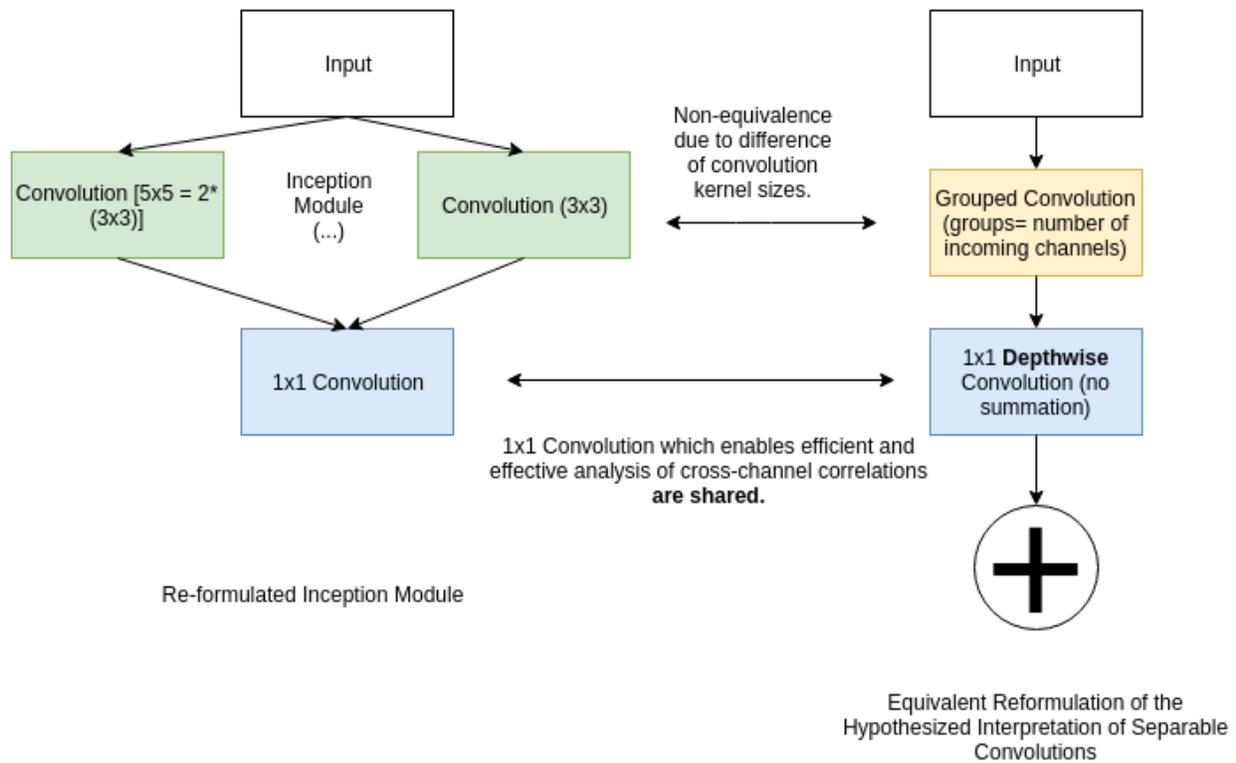

Figure 3

# 4. Experimental Justification

In order to provide experimental justification for our hypothesis, we compare several hypothesized interpretations of separable convolutions to separable convolutions on the well-known CIFAR-10 dataset. We compare several competing hypotheses to the baseline network of stacked separable convolutions. We wait until they finish training at 190 epochs and then compare their performance on a validation/test set of 6000 randomly selected images. We then take the absolute value of the difference in their error percentages. By averaging across 10 trials, the most accurate formulation should provide us with the least absolute difference from test error.

**Absolute Difference in Test Error on CIFAR-10 Averaged Across 10 Trials**

| Setup | Absolute difference from baseline |
| --- | --- |
| Baseline- Separable Convolution Network | 0 |
| Inception Modules | 0.83% |
| ResNeXt Modules | 0.62% |
| Re-formulated Inception Modules | 0.71% |
| Separable Convolution Equivalent Reformulation (center in Figure 1) (Strict equivalence, simply used as a quasi-control to discern the impact of random deviation) | 0.08% |
| Hypothesized Interpretation of Separable Convolutions (left in Figure 2) A.k.a- "The Hybrid Hypothesis" | 0.12% |

Figure 4

Another piece of experimental evidence which supports this interpretation comes from Chollet's XCeption paper in 2016, which showed that adding a nonlinearity between the depthwise convolution and pointwise convolution actually degrades accuracy. If the hybrid hypothesis is true, this makes sense, since nonlinearities (particularly ReLU) would destroy the integrated module structure of the layer and instead separate it into a simple multilayer network.

As an aside, we notice that this interpretation seems to suggests that separable convolutions would not perform well in a setup such as FractalNet (Larsson et al, 2016) which has its own module structure. Indeed, this is the case as we show that replacing FractalNet's convolution layers with separable convolution layers has little to no impact upon performance, whereas by contrast, replacing DarkNet's (Redmon et al, 2015) layers with separable convolutions leads to tremendous performance improvements. However, we also note that only one trial was done and therefore these particular results should be taken with caution.

**FractalNet with Separable Convolutions vs Without on CIFAR-10**

| Separable Convolutions | Final Accuracy (in test error %) |
| --- | --- |
| Yes | 6.15% |
| No | 7.61% |

Figure 5

**DarkNet with Separable Convolutions vs Without on CIFAR-10**

| Separable Convolutions | Final Accuracy (in test error %) |
|---|---|
| Yes | 6.03% |
| No | 11.9% |

Figure 6

# 5. Conclusion

In this work we proposed a new interpretation of separable convolutions that demonstrates that they can be interpreted as a combination of ResNeXt modules and Inception modules. We also reinforced this interpretation with experimental results and showed that experimental results matched with the predictions from our hybrid hypothesis. We believe that this new interpretation of separable convolutions will lead to better utilization of separable convolutions in future deep convolutional architectures. Now that their underlying mechanism of efficacy is better understood, it can be better exploited by network architects and therefore lead to better performance on major datasets and competitions. This could potentially usher in the era of leveraging separable convolutions as a means to increase accuracy rather than accelerate inference. Finally, we note that separable convolutions also enjoy faster runtime performance, which will help immensely with the mainstream deployment of deep learning.

# References/Bibliography